\documentclass[conference]{IEEEtran}
\IEEEoverridecommandlockouts
\usepackage{amsmath,amssymb,amsfonts}
\usepackage{algorithmic}
\usepackage{graphicx}
\usepackage{textcomp}
\usepackage{xcolor}
\usepackage{pifont}
\usepackage{booktabs}
\usepackage{subcaption}
\usepackage[backend=biber,style=ieee]{biblatex}
\usepackage{float}
\usepackage{fancyhdr}

\fancypagestyle{preprintfirst}{%
    \fancyhf{}
    \fancyhead[L]{\normalfont\normalsize Submitted to IEEE PES International Meetings 2027}

}

\begin{document}

% TODO: confirming title
\title{
% Assessing the Impact of Graph Representation on L2RPN Performance
A Comparative Study of Graph Representations for GNN-Based Power Grid Control in L2RPN
% Graph Representations for Reinforcement Learning-Based Topology Control in L2RPN
% A Controlled Comparison of Graph Representations for GNN-Based Topology Control in L2RPN
}

\author{\IEEEauthorblockN{Adrian Degenkolb\IEEEauthorrefmark{1}}
\IEEEauthorblockA{
    % \textit{Institute for Automation and Applied Informatics} \\
\textit{Karlsruhe Institute of Technology}\\
Karlsruhe, Germany \\
adrian.degenkolb@partner.kit.edu} % ordic: 0009-0002-7578-5669
\and
\IEEEauthorblockN{Qiong Huang\IEEEauthorrefmark{1}\IEEEauthorrefmark{2}}
\IEEEauthorblockA{
    % \textit{Institute for Automation and Applied Informatics} \\
\textit{Karlsruhe Institute of Technology}\\
Karlsruhe, Germany \\
qiong.huang@kit.edu} % orcid: 0000-0002-1958-6094
\and
\IEEEauthorblockN{Benjamin Sch\"{a}fer}
\IEEEauthorblockA{
    % \textit{Institute for Automation and Applied Informatics} \\
\textit{Karlsruhe Institute of Technology}\\
Karlsruhe, Germany \\
benjamin.schaefer@kit.edu} % orcid: 0000-0003-1607-9748

\thanks{\IEEEauthorrefmark{1} These authors contributed equally to this work.}
\thanks{\IEEEauthorrefmark{2} Corresponding author.}
}

\maketitle
\thispagestyle{preprintfirst}

\begin{abstract}
Graph construction is a critical but underexamined design choice in deep reinforcement learning for power grid control. 
We present a controlled experimental comparison of different graph representations, including physical topology, electrical-sensitivity, and hybrid variants for topology control in the Learning to Run a Power Network (L2RPN) environment. 
%All agents use an identical GNN-PPO architecture trained over five independent seeds. Results show that a compact substation-level graph achieves the highest test survival (99.3\% steps) with the lowest variance, while richer element-level and heterogeneous representations yield no systematic advantage. Electrical-sensitivity graphs provide no improvement over physical topology alone. Policy spatial patterns are qualitatively similar across methods, indicating that graph representation affects learning efficiency and robustness more than the discovered control strategy. 
%These findings argue for matching graph complexity to task granularity rather than maximizing representational richness, and highlight the importance of controlled representation studies at scale.
Our findings indicate that matching graph complexity to task granularity is more important than maximizing representational richness, and highlight the importance of controlled representation studies at scale.
\end{abstract}
% Graph construction is not a neutral preprocessing choice. In a controlled L2RPN study, compact topology-based representations provide stronger and more consistent grid-control performance than richer element-level or fully connected electrical-sensitivity representations.


\section{Introduction}
\label{sec:introduction}
% 1 page maximum
Variable renewable generation increases fluctuations in power injections and can aggravate transmission congestion \cite{dorfer2022power}. Grid operators commonly mitigate congestion through measures such as redispatch and curtailment, which can be costly and counteract the efficient use of renewable energy. Topology reconfiguration offers an alternative by switching lines or changing busbar assignments to redirect power flows \cite{van2025optimizing}. Selecting these topology actions is challenging, as their effects are nonlocal and sequential, while the number of feasible configurations grows rapidly with grid size.
% Its operational potential comes with a difficult decision problem: topology actions have nonlocal effects, their feasible combinations grow rapidly with grid size, and a beneficial immediate action may reduce security later in an operating sequence.

Reinforcement learning (RL) addresses such sequential decisions by learning a policy through interactions with the environment. 
% In power-grid control, an agent observes the current network state, selects a control action, and receives feedback that reflects secure and sustained operation. 
Learning to Run a Power Network (L2RPN) standardizes this setting through the package Grid2Op \cite{marot2020learning,grid2op}. Its environments combine time-varying generation and demand with line limits, contingencies, cooldown constraints, and cascading disconnections. An agent must therefore maintain secure operation over long episodes rather than solve a single static operating point. L2RPN competitions and subsequent studies have demonstrated the potential of RL while highlighting challenges in action-space design, safety, and generalization \cite{van2025optimizing}.

Graph neural networks (GNNs) encode grid structure by sharing parameters across components and propagating information along edges \cite{scarselli2008graph,ringsquandl2021power}. However, the physical power system does not prescribe a unique computational graph. Nodes may represent substations, busbars, line endpoints, or individual elements, while edges may encode physical, switchable, or electrical relations. This choice determines the state granularity, message-passing paths, and computational cost.

Graph construction is nevertheless rarely isolated from choice of RL algorithms, action space, and policy architecture \cite{van2025optimizing,hassouna2026graph}. We address this gap by comparing topology-based, power-flow-based, and hybrid graph representations in \texttt{l2rpn\_case14\_sandbox} under a fixed GNN-PPO pipeline with action space, training budget, evaluation set, and five random seeds. A non-graph MLP serves as an additional baseline. We compare learning behavior, held-out survival, and spatial action patterns. Our results show that graph representation materially affects learning and robustness, with the compact substation graph outperforming more detailed physical and electrical-sensitivity representations.

% The remainder of this paper is organized as follows. Section~\ref{sec:related_work} reviews related work on L2RPN and graph representations. Section~\ref{sec:methodology} describes the graph representations and GNN architectures. Section~\ref{sec:experiment_results}  presents the experimental setup, learning curves, test performance, and spatial action patterns. Section~\ref{sec:conclusions} concludes with a discussion of implications and future directions. 
Code and supplementary material are available in our public repository\footnote{\url{https://github.com/KIT-IAI-DRACOS/L2RPNGraphReprComparison}}.

\section{Related Work}
\label{sec:related_work}
\textbf{RL-based topology control:} 
The combinatorial L2RPN action space has motivated restricted action sets and action-selection schemes \cite{subramanian2021exploring,zhou2021action}, semi-Markov actor-critic policies \cite{yoon2021winning}, and AlphaZero-inspired planning \cite{dorfer2022power,zetto2026learning}. Hierarchical and multi-agent approaches further decompose decisions across control levels or grid regions \cite{manczak2023hierarchical,van2023multi}. Although these methods demonstrate effective topology control, simultaneous differences in agent architecture, actions, and evaluation prevent their results from isolating the observation representation \cite{van2025optimizing}.

\textbf{Graph learning for power systems:} GNNs have been applied to fault localization, Volt-Var control, and transmission-line flow control \cite{chen2019fault,lee2022graph,xu2020simulation}. Their shared local computations naturally accommodate changing network connectivity, including topology variation \cite{taha2022learning}. Yet using a GNN does not determine the mapping from physical components to nodes and edges; alternative mappings induce different neighborhoods even for the same grid state.
% GNNs introduce a relational inductive bias by applying shared message-passing operations to network components and their connections. This structure has been explored across power-system applications because it mirrors, at least approximately, the sparse interactions imposed by electrical networks \cite{ringsquandl2021power}. Examples include graph-convolutional fault localization \cite{chen2019fault}, graph policies for Volt-Var control \cite{lee2022graph}, and graph RL for transmission-line flow control \cite{xu2020simulation}. In topology-control settings, a graph encoder can adapt its connectivity to switching actions and can reuse the same local computations across buses or substations. Taha \emph{et al.}, for example, study learning under varying grid topology \cite{taha2022learning}. Nevertheless, using a GNN does not determine how the physical system should be mapped to nodes and edges. Different mappings induce different neighborhoods and message-passing distances even when they encode the same underlying grid state.

\textbf{Graph representations in L2RPN:} Most graph-based L2RPN agents use the default Grid2Op construction based on line endpoints and bus connections \cite{grid2op,hassouna2026graph}. De Jong \emph{et al.} retain these nodes but distinguish line, active same-bus, and switchable cross-bus relations by edge types \cite{de2025generalizable}. Batanero \emph{et al.} instead propose a detailed Element graph containing individual equipment and feasible connections \cite{batanero2025graph}. These studies establish several viable physical abstractions, but differing learning and evaluation settings preclude a controlled comparison.
% Their substation graph represents active busbars and connects them by power lines, producing a compact view of the current network, whereas their element graph represents generators, loads, lines, storage units, and busbars individually and encodes both active and feasible connections. These proposals demonstrate that multiple physical graph abstractions can support topology control. However, they were introduced with different model and evaluation choices, so their reported results do not isolate representation quality.

% Position of this work 
% Existing surveys identify observation design and graph construction as relevant but comparatively underexplored choices in RL-based topology optimization \cite{van2025optimizing,hassouna2026graph}. 
Prior L2RPN graphs are predominantly topology based, although physical adjacency is only one description of grid interaction. Power Transfer Distribution Factors (PTDF) describe how injections affect line flows \cite{evrenosoglu2004effects}, Line Outage Distribution Factors (LODF) characterize outage-induced flow redistribution \cite{guler2007generalized}, and the bus-impedance matrix captures self and transfer impedances \cite{aldaoudeyeh2019modeling}. We compare graphs derived from these quantities and with the physical constructions above and their hybrids while holding the learning and evaluation pipeline fixed.
% GNN, PPO algorithm, action space, training data, and evaluation procedure fixed, our study attributes observed performance differences specifically to graph representation.

\section{Methodology}
\label{sec:methodology}                             
                                                           
We evaluate three categories of graph representations. \textbf{Topology-based} representations define edges from physical grid connectivity. 
\textbf{Power-flow-based} representations use fully connected graphs whose edge features encode electrical sensitivities. \textbf{Hybrid representations} combine physical and electrical relations and distinguish them by edge types. The node features used by each representation are provided in the supplementary material.
% \textcolor{red}{Github}.
% TODO - move appendix as supplentary material in GitHub - file Supplementary
% Table~\ref{tab:node_features}.

\subsection{Topology-Based Representations}
\begin{figure*}[htbp]
  \centering
  \includegraphics[width=0.9\linewidth]{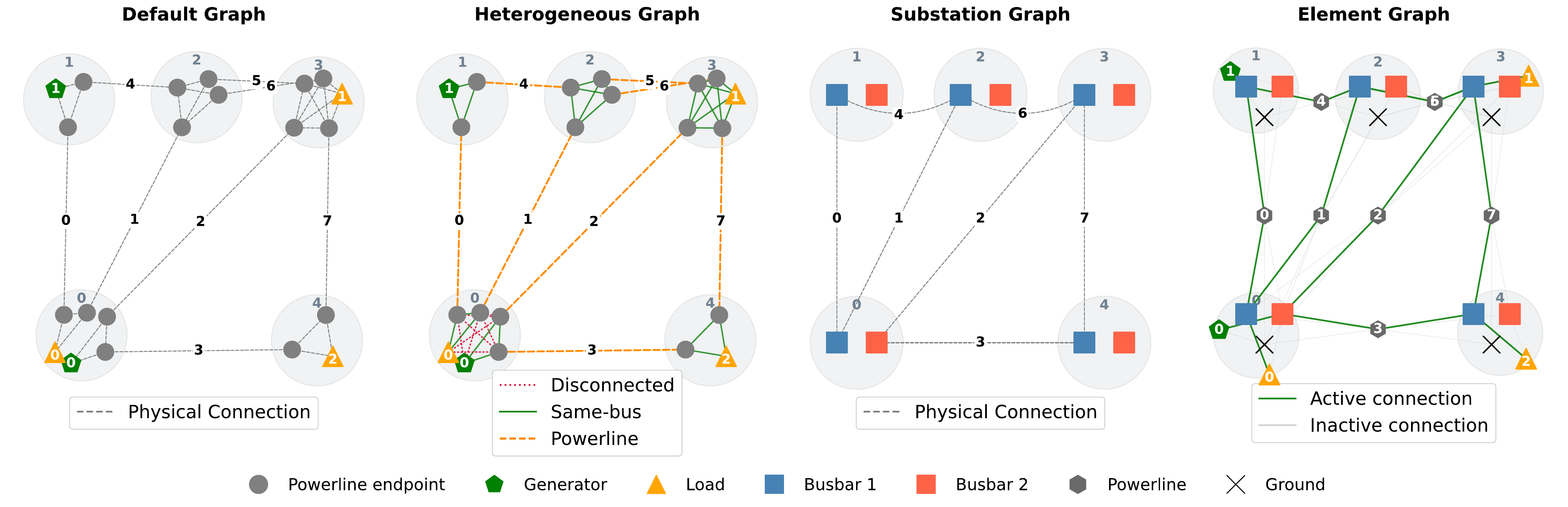}
  \caption{Topology based representations of the
\textit{rte\_case5\_example} environment with 5 substations (0-4). Each panel depicts the same grid configuration. All elements are assigned to Busbar 1 except at substation 0 (bottom left), where several elements are assigned to Busbar 2.}
% TODO: check the captions of this figure
% in which everything is connected to the first bus with the exception of substation 0 (bottom left substation), which connects several elements to the second bus.}
  \label{fig:topology_based_representations}
\end{figure*}

Figure~\ref{fig:topology_based_representations} shows the four topology-based representations.

\paragraph{Default Graph} follows the graph construction provided by \cite{grid2op} and is the representation most commonly adopted in prior L2RPN studies \cite{hassouna2026graph}. Nodes represent power line endpoints and element-to-bus connections. An edge connects two nodes when they are the endpoints of the same power line or are assigned to the same substation and bus.

\paragraph{Heterogeneous Graph} proposed by \cite{de2025generalizable}, uses the same nodes as the default graph but distinguishes three edge types: power line edges, same-busbar connections within a substation, and inactive cross-busbar connections that can be activated through switching. The third edge type explicitly represents feasible topology changes that are absent from the current physical configuration.
% same-bus edges (active physical connections), and same-substation cross-bus edges (currently inactive but switchable connections).

\paragraph{Substation Graph} proposed by \cite{batanero2025graph}. Each node represents an active busbar and each edge represents a power line between them. This construction provides a compact representation of the current physical topology.

\paragraph{Element Graph} also introduced by \cite{batanero2025graph}, assigns a node to every generator, load, power line, storage unit, and busbar. Each substation additionally contains a virtual ground busbar to which disconnected elements are assigned. Edges connect every physically connectable pair of nodes, and a binary edge feature indicates whether the corresponding connection is currently active.

% \subsection{Power-Flow-Based Representations}
% \begin{figure*}
%   \centering
%   \includegraphics[width=0.9\linewidth]{figures/powerflow_and_combined_representations.png}
%   \caption{Power-Flow-Based (left) and hybrid (right)
% representations of the \textit{rte\_case5\_example} environment.}
% \label{fig:powerflow_based_and_combined_representations} 
% \end{figure*}

\subsection{Power-Flow-Based Representations}
\begin{figure}[htbp]
  \centering
  \includegraphics[width=\linewidth]{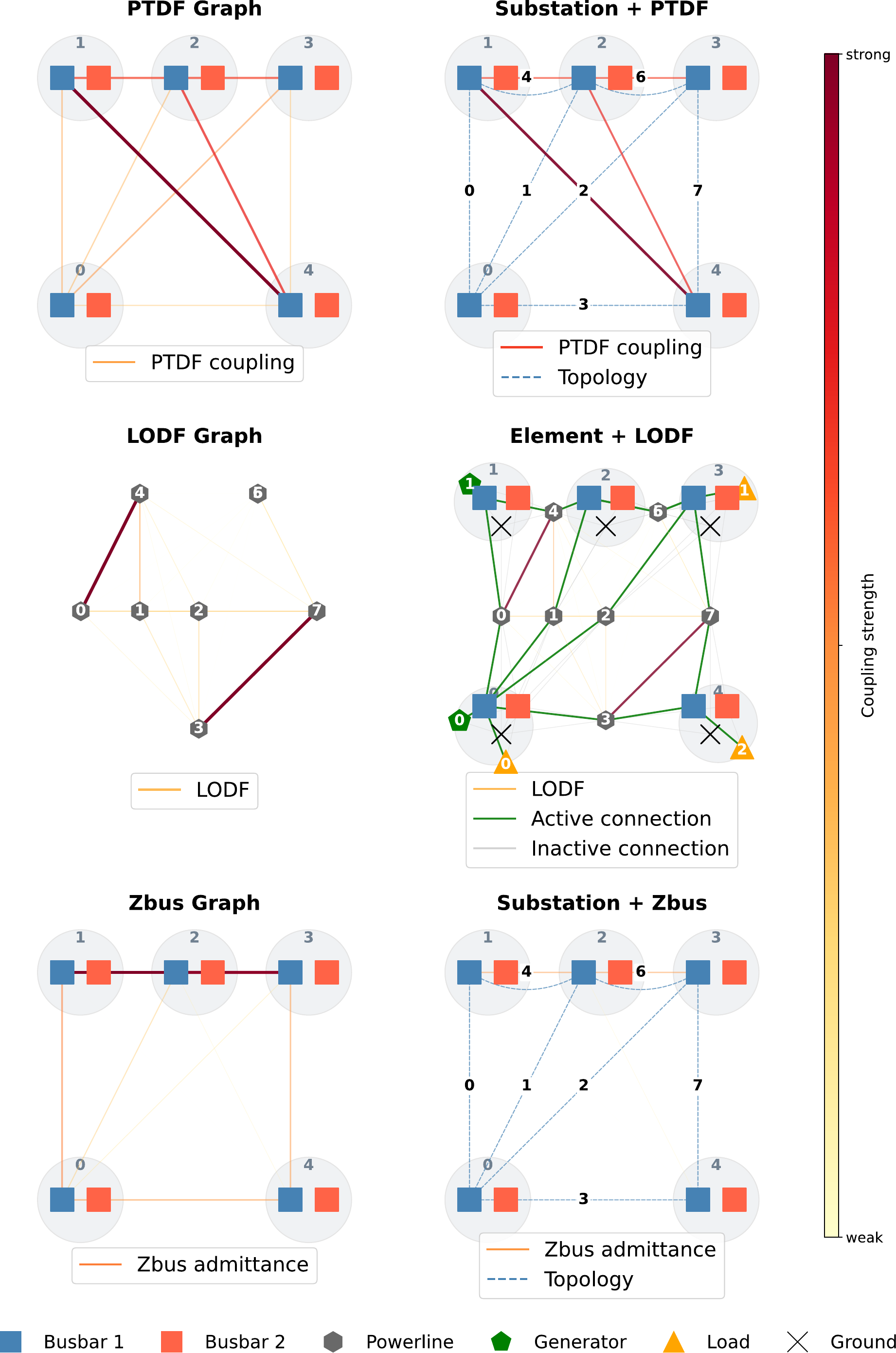}%\includegraphics{figures/powerflow_and_combined_representations.svg}
  \caption{Power-Flow-Based (left) and Hybrid (right) representations of the \textit{rte\_case5\_example} environment.}
\label{fig:powerflow_based_and_combined_representations}
\end{figure}

Figure~\ref{fig:powerflow_based_and_combined_representations} shows the power-flow-based and the hybrid representations that are fully connected directed graphs.
Every edge carries a scalar feature that quantifies an electrical relationship between its incident nodes under a DC power-flow approximation. The node definition and the meaning of the edge feature depend on the selected electrical quantity.

\paragraph{PTDF Graph} each node represents an active busbar. For busbar $i$ and $j$, the edge feature is the PTDF-based distance
\begin{equation}
  D[i,j] = \sum_l \left|\mathrm{PTDF}[l,i] - \mathrm{PTDF}[l,j]\right|, 
\end{equation}
where $\mathrm{PTDF}[l,i]$ quantifies the sensitivity of the active-power flow on line $l$ to an injection at busbar $i$ under the selected convention \cite{evrenosoglu2004effects}. A large $D[i,j]$ indicates that injections at the two busbars have substantially different effects on network flows.     

\paragraph{LODF Graph} each node represents a power line. The feature of the directed edge from line $j$ to line $i$ is $|\mathrm{LODF}_{ij}|$, where $\mathrm{LODF}_{ij}$ denotes the fraction of the pre-outage flow on line $j$ that is transferred to line $i$ following the outage of line $j$ \cite{guler2007generalized}. Taking the absolute value retains the magnitude of the contingency coupling while discarding its direction.
% The edge feature $|\text{LODF}[i,j]|$ is the Line Outage Distribution Factor: the fraction of line $j$'s pre-fault flow that shifts to line $i$ if $j$ trips, quantifyingcontingency coupling between lines.
              
\paragraph{ZBus Graph} uses the same active-busbar nodes as the PTDF graph. Its edge feature is an impedance-derived coupling weight
% \begin{equation}
  $w_{ij} = \frac{1}{|Z_{ij}| + 1}$,
% \end{equation}
where $Z_{ij}$ is an off-diagonal transfer-impedance entry of $Z_\mathrm{bus} = \mathrm{pinv}(Y_\mathrm{bus})$ and $\mathrm{pinv}$ denotes the Moore--Penrose pseudo-inverse, which yields the minimum-norm solution to $V = Z_{\mathrm{bus}} I$ when $Y_{\mathrm{bus}}$ is singular. Low impedance indicates strong electrical coupling between buses.                
              
\subsection{Hybrid representations}
Each hybrid representation combines a topology-based graph with a power-flow-based graph and assigns the two relations distinct edge types. \textbf{Element + LODF} combines the element graph (edge type~0) with fully-connected LODF edges between power-line nodes (edge type~1). \textbf{Substation + PTDF} combines the substation graph (edge type~0) with PTDF-distance edges (edge type~1). \textbf{Substation + ZBus} has the same structure but replaces the PTDF distances with the ZBus coupling weights.

\subsection{GNN Architecture}
All graph-based agents use the same GNN encoder. Relative to a standard graph convolution \cite{kipf2016semi}, the encoder should accommodate both multiple edge types and scalar edge features. We therefore associate each edge type with a separate transformation matrix and map scalar edge features to message weights. Given node features $\mathbf{x}_v$, the encoder is defined by:
% handle both in a single formulation. Starting from projected node features, each message-passing layer proceeds as:
\begin{align}
  h_v^{(0)}     &= \mathrm{BN}(f(\mathbf{x}_v))
\tag{node projection} \\[3pt]                            
  w_{uv}^{(k)}  &=
\mathrm{MLP}_k\!\left(e_{uv}^{(k)}\right) \tag{edge      
weights} \\[3pt]
  \hat{h}_v^{(l)} &= \sum_{k=0}^{K-1}\ \sum_{u \in     
\mathcal{N}_k(v) \cup \{v\}}                             
                   \frac{w_{uv}^{(k)}}{\sqrt{\tilde{d}_
{v,k}}\,\sqrt{\tilde{d}_{u,k}}}\,                        
                   W_k^{(l)}\, h_u^{(l)}
\tag{aggregation} \\[3pt]                                
  h_v^{(l+1)}   &= \mathrm{ELU}\!\left(\mathrm{BN}(\hat{h}_v^{(l)})\right) + h_v^{(l)} \tag{update},
\end{align}
where $f$ denotes the input projection, $\mathrm{BN}$ denotes batch normalization, $K$ is the number of edge types, $\mathcal{N}_k(v)$ is the neighborhood of node $v$ under edge type $k$, and $W_k^{(l)}$ is the
corresponding transformation weight matrix at layer $l$. $\tilde{d}_{v,k}$ is the degree of $v$ after adding self-loops, and $e_{uv}^{(k)}$ is the   
scalar feature of edge $(u,v)$ of type $k$. When a representation has no scalar edge features, we set $w_{uv}^{(k)}=1$. After $L$ message-passing layers, the graph-level embedding $o = \mathrm{MeanPool}(f'(\{h_v^{(L)}\}))$ is passed to the same MLP used for the non-graph baseline.

\section{Experiments and Results}  
\label{sec:experiment_results}
\subsection{Experimental Setup}
We evaluate the ten graph representations and a non-graph MLP baseline using five independent training seeds, yielding 55 trained runs in total. The MLP receives the grid states as a flat vector, whereas the graph-based agents process the corresponding structured observations with the GNN encoder described in Section~\ref{sec:methodology}. All agents are trained with Proximal Policy Optimization (PPO) \cite{schulman2017proximal} using RLlib \cite{liang2018rllib}. Our implementation builds on the experimental pipeline of \cite{van2025optimizing}. The GNN architecture and PPO hyperparameters are held fixed across graph-based methods; only the graph converter is changed. Full architecture and hyperparameter settings are reported in the supplementary material.
% \textcolor{red}{GitHub}.
% Table \ref{tab:gnn-architecture} and \ref{tab:ppo-hyperparameters}.

Training is conducted in \texttt{l2rpn\_case14\_sandbox}, a Grid2Op environment of the IEEE 14-bus system. The action space comprises the topology changes proposed by \cite{subramanian2021exploring}. Evaluation uses a fixed set of episodes whose chronics are excluded from training. We report two complementary metrics: \emph{Surv.\ Episodes}, the percentage of evaluation episodes completed, and (\emph{Surv.\ Steps}), the percentage of all available evaluation steps completed. Reported values are the mean and standard deviation across five seeds, except for Substation + ZBus, which uses the three successful seeds.
% We also report the mean and standard deviation across the five training seeds.
In addition to held-out performance, we record rollout survival during training to compare learning speed and stability. 
% curves of survival duration to assess convergence behavior. 
% We further analyze agent behavior through substation action frequency profiles and spatial congestion maps, aggregated over test episodes.

% \section{Results}
% \label{sec:results}
\subsection{Performance Comparison}
\begin{figure}[htbp]
    \centering
    \begin{subfigure}[t]{\linewidth}
        \centering
        \includegraphics[width=\linewidth]{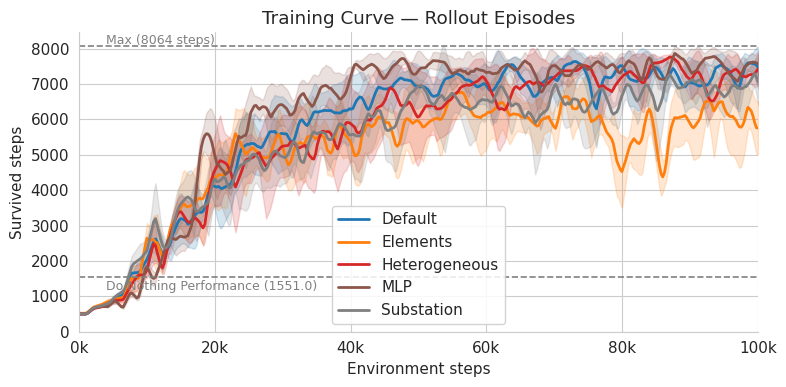}
        % \caption{Rollout survival during training for the topology-based representations and the MLP baseline.}
        \label{fig:perf_topology_based}
    \end{subfigure}
% \end{figure}
    % \hfill
    \vspace{-1em}
    \begin{subfigure}{\linewidth}
% \begin{figure}[htbp]
        \centering
        \includegraphics[width=\linewidth]{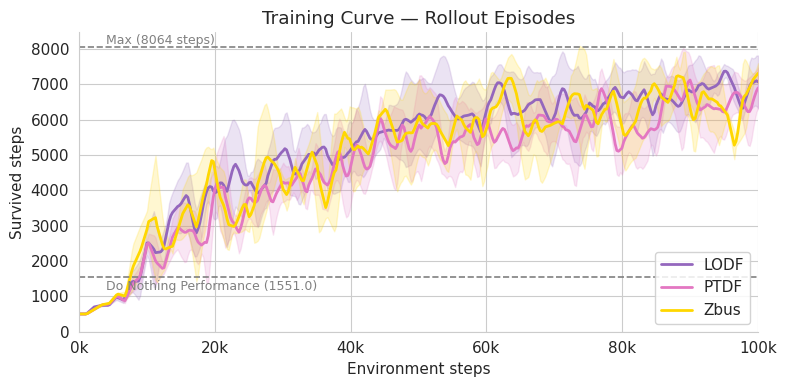}
        % \caption{Rollout survival during training time of power-flow-based graph representations.} 
        \label{fig:perf_powerflow_based}
    \end{subfigure}
    \vspace{-1em}
    % \caption{Survival duration vs training time of topology based graph representations (left) and powerflow based graph representations (right).}
    % \label{fig:performance_comparison_topo_and_power} 
% \end{figure}
\begin{subfigure}{\linewidth} %[htbp]
    \centering
    \includegraphics[width=\linewidth]{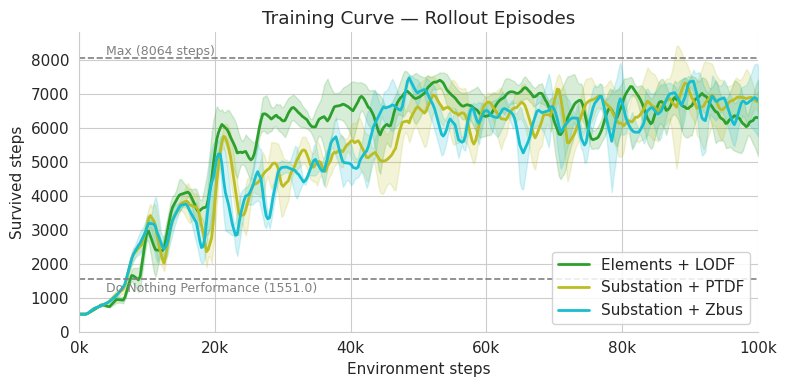}
    % \caption{Rollout survival during vs training for hybrid graph representations.}
    \label{fig:performance_comparison_combined}
    \end{subfigure}
    \caption{Rollout survival during training for the topology-based graph and the MLP baseline (top), powerflow-based graphs (middle), and hybrid graphs (bottom).}
    \label{fig:compare_all}
\end{figure}

% Figures~\ref{fig:perf_topology_based} and~\ref{fig:perf_powerflow_based} 
Figure~\ref{fig:compare_all} (top and middle) shows rollout survival during training. All methods improve substantially beyond the do-nothing reference. Among topology-based methods, the default graph converges the fastest while the Element graph learns more slowly and exhibits larger late-training declines. Electrical-sensitivity graphs generally remain below the strongest topology-based methods. Among the hybrids, Element + LODF improves fastest initially, while the two Substation hybrids later reach similar training survival (see Figure~\ref{fig:compare_all} (bottom)).
% Figure~\ref{fig:performance_comparison_combined}).
% The power-flow-based representations follow broadly similar trajectories but generally remain below the strongest topology-based methods. Figure~\ref{fig:performance_comparison_combined} shows hybrid representations: Element + LODF improves most rapidly early in training, while Substation + PTDF and Substation + ZBus reach comparable survival levels later in training.

Held-out results in Table~\ref{tab:performance} favor compact physical representations. Substation achieves the highest survival ($96.7 \pm 1.9\%$ of episodes and $99.3\pm0.5\%$ of steps), followed by Default in episode survival. LODF is the strongest standalone electrical representation, while the MLP exhibits the greatest variability across seeds. Adding LODF to Element raises episode survival from $86.7\%$ to $92.4\%$ and step survival from $94.3\%$ to $95.9\%$, whereas neither PTDF nor ZBus improves the Substation graph. Thus, electrical edges can compensate for a weak physical abstraction but provide no consistent benefit to the compact Substation representation.
% competitive at $94.8\pm1.8\%$ episode survival, whereas PTDF and ZBus attain $90.0\pm5.5\%$ and $88.1\pm3.4\%$, respectively. The MLP baseline lies near the middle of the ranking and shows the largest variability across seeds in both metrics.
% Power-flow-based representations cluster in the lower half. The MLP baseline places in the middle with the highest variance ($\pm 6.4\%$).

%Figure~\ref{fig:performance_powerflow} compares each power-flow-based representation against its hybrid counterpart. 
% Adding LODF edges to the Element graph increases mean episode survival from $86.7\%$ to $92.4\%$ and mean step survival from $94.3\%$ to $95.9\%$. 
% a clear improvement: the elements graph is the weakest representation overall, and the additional electrical connectivity partially compensates for this. 
% In contrast, neither electrical augmentation improves on the Substation graph: Substation + PTDF and Substation + ZBus reach mean episode-survival rates of $93.3\%$ and $95.2\%$, compared with $96.7\%$ for the topology-only Substation graph. Thus, electrical edges can compensate for a weak physical abstraction, as in Element + LODF, but provide no consistent benefit when added to the compact Substation representation.

% Adding PTDF or ZBus edges to the substation graph, however, produces no significant change, suggesting the substation topology already captures the relational structure relevant to this environment.

% Despite differences in convergence speed and test performance, the learned policies exhibit qualitatively similar spatial behavior. 
Across representations, actions concentrate on substations 3, 4, and 8, with recurring congestion in the 4–5–12 and the 3–8–6–7 regions (see the supplementary material). This similarity suggests that graph construction affects learning efficiency and robustness more than the selected control regions.
% \textcolor{red}{GitHub}).
% (see Figure~\ref{fig:all} and ~\ref{fig:behaviour}).
% This similarity suggests that graph construction primarily affects how efficiently and robustly the policy is learned rather than which grid regions are ultimately controlled.

\begin{table}[h!]
    \centering
    \caption{Performance comparison of different graph representations.}
    \label{tab:performance}
    \begin{tabular}{lcc}
        \toprule
        Method
            & Surv. Episodes (\%)
            & Surv. Steps (\%) \\
        \midrule
        Substation
            & $96.7 \pm 1.9$
            & $99.3 \pm 0.5$ \\
        Default
            & $96.2 \pm 1.2$
            & $97.6 \pm 0.7$ \\
        Substation + ZBus$^\dagger$
            & $95.2 \pm 0.0$
            & $97.5 \pm 0.0$ \\
        LODF
            & $94.8 \pm 1.8$
            & $97.3 \pm 1.2$ \\
        Heterogeneous
            & $93.3 \pm 1.8$
            & $96.8 \pm 0.8$ \\
        Substation + PTDF
            & $93.3 \pm 2.8$
            & $96.5 \pm 1.0$ \\
        MLP
            & $92.9 \pm 6.4$
            & $96.3 \pm 3.5$ \\
        Element + LODF
            & $92.4 \pm 3.5$
            & $95.9 \pm 2.0$ \\
        PTDF
            & $90.0 \pm 5.5$
            & $94.9 \pm 2.8$ \\
        ZBus
            & $88.1 \pm 3.4$
            & $94.4 \pm 1.4$ \\
        Element
            & $86.7 \pm 4.2$
            & $94.3 \pm 2.4$ \\
        \bottomrule
    \end{tabular}\par
    \vspace{0.2em}
    \parbox{0.98\linewidth}{%
        \raggedright\footnotesize
        $^\dagger$ Based on three successful seeds; two runs failed numerically; step-survival standard deviation rounds to $0.0$.}
\end{table}

% \parbox{\linewidth}{\footnotesize
% $^\star$ Mean $\pm$ standard deviation over three successful runs; two runs terminated because of numerical instability. The step-survival standard deviation is nonzero but below $0.05$ percentage points and therefore rounds to $0.0$.}

%\begin{figure}
%    \centering

%    \begin{subfigure}[t]{\linewidth}
%        \centering
%        \includegraphics[width=\linewidth]{figures/performance_lodf.png}
%        \label{fig:performance_lodf}
%    \end{subfigure}

%    \vspace{0.5em}

%    \begin{subfigure}[t]{\linewidth}
%        \centering
%        \includegraphics[width=\linewidth]{figures/performance_zbus.png}
%        \label{fig:performance_zbus}
%    \end{subfigure}

%    \vspace{0.5em}

%    \begin{subfigure}[t]{\linewidth}
%        \centering
%        \includegraphics[width=\linewidth]{figures/performance_ptdf.png}
%        \label{fig:performance_ptdf}
%    \end{subfigure}

%    \caption{Performance of the powerflow-based graph representations.}
%    \label{fig:performance_powerflow}
%\end{figure}

\section{Conclusions}
\label{sec:conclusions}                                                       
We compared ten graph representations and a non-graph MLP baseline under a fixed GNN-PPO pipeline. Every representation supported a viable policy, but learning behavior and held-out survival varied substantially. The compact Substation graph achieved the highest mean survival, closely followed by the Default, while richer physical and 
% the more detailed Heterogeneous and Element graphs provided no systematic advantage.
standalone PTDF- and ZBus-based graphs offered no systematic advantage. 
% likewise did not improve on the strongest physical representations. information was nevertheless useful in specific settings: the standalone 
LODF graph remained competitive, and augmenting the Element graph when used in a hybrid form.
% with LODF edges improved its mean survival. Graph construction should therefore be treated as a consequential modeling choice rather than a neutral preprocessing step, while greater representational detail should not be assumed to yield better control.
% Currently in L2RPN literature the \textit{default graph} is by far the most popular \cite{hassouna2026graph} and only few alternatives have been explored by \cite{de2025generalizable, batanero2025graph}. Therefore we argue, that future work should further focus on establishing and benchmarking novel graph representations and GNN architectures to process them. 

These results establish graph construction as a consequential design choice
but do not support maximizing representational detail. Because the study is
limited to one benchmark, one action set, and one RL architecture, 
% The learned policies exhibited similar spatial action and congestion patterns across representations, suggesting that graph construction influenced learning efficiency and robustness more strongly than the qualitative control strategy. These conclusions are limited to one small benchmark network, one action set, and one RL architecture. Moreover, the fully connected electrical graphs require a number of edges that grows quadratically with the node count, which restricts their direct application to larger systems. 
future work should evaluate larger and structurally different L2RPN environments, as well as examine sparse or learned electrical-sensitivity graphs to identify representations that remain effective and computationally practical at operationally relevant scales.
%Future work should evaluate these findings across larger and structurally different L2RPN environments, examine sparse or learned electrical-sensitivity graphs, and study interactions between graph construction and alternative GNN architectures. Such comparisons are necessary to determine which representations remain effective and computationally practical at operationally relevant scales.
% Although topology-based representations outperform power-flow-based ones, we show that physical connectivity is not the only working foundation for connectivity in graph representations. For example, the LODF graph ranks 4th on test episodes and improves the performance of the elements graph when layered into a combined graph representation. Importantly we do not suggest that the power flow based representations introduced by us should be applied in practical use since they introduce fully-connected edges that scale quadratically in the number of nodes, limiting applicability to larger grids. Instead, we highlight that the variety of working methods is larger than what is suggested by the limited set of representations used in prior work.

\section*{Acknowledgment}
% funding, HPC usage 
This work is supported by the Helmholtz Association under grant no. VH-NG-1727 and through Helmholtz AI. Computing time was provided by NHR@KIT on HoreKa, jointly supported by the responsible federal and Baden-W{\"u}rttemberg ministries and partly funded by DFG. ChatGPT was used for language refinement and Claude for coding assistance. All technical content and interpretations were developed and verified by the authors.

%% reference 
\printbibliography
% \bibliographystyle{IEEEtran}
% \bibliography{reference}

\end{document}